\documentclass[conference]{IEEEtran}
\IEEEoverridecommandlockouts
\usepackage{cite}
\usepackage{amsmath,amssymb,amsfonts}
\usepackage{algorithmic}
\usepackage{graphicx}
\usepackage{textcomp}
\usepackage{xcolor}
\usepackage{tabularx,booktabs}
\newcolumntype{C}{>{\centering\arraybackslash}X} 
\usepackage{lipsum} % filler text

\def\BibTeX{{\rm B\kern-.05em{\sc i\kern-.025em b}\kern-.08em
    T\kern-.1667em\lower.7ex\hbox{E}\kern-.125emX}}
\begin{document}

\title{Enhancing Low Resource NER Using Assisting Language And Transfer Learning
}

\author{Maithili Sabane$^{\hspace{4pt}*1,3}$, Aparna Ranade $^{ *1,3}$, Onkar Litake$^{\hspace{4pt}*1,3}$, Parth Patil $^{ *1,3}$, \\
Raviraj Joshi$^{\hspace{4pt}*2,3}$, Dipali Kadam $^{1}$ \\

$^{1}$Pune Institute of Computer Technology, Pune\\
$^{2}$Indian Institute of Technology Madras, Chennai\\
$^{3}$ L3Cube, Pune\\

\texttt{\{msabane12,aparna.ar217,litakeonkar,parthpatil8399\}@gmail.com}\\ 
\texttt{ravirajoshi@gmail.com,}
\texttt{ddkadam@pict.edu}
}

\maketitle

\begin{abstract}

 Named Entity Recognition (NER) is a fundamental task in NLP that is used to locate the key information in text and is primarily applied in conversational and search systems. In commercial applications, NER or comparable slot-filling methods have been widely deployed for popular languages. NER is used in applications such as human resources, customer service, search engines, content classification, and academia. In this paper, we draw focus on identifying name entities for low-resource Indian languages that are closely related, like Hindi and Marathi. We use various adaptations of BERT such as baseBERT, AlBERT, and RoBERTa to train a supervised NER model. We also compare multilingual models with monolingual models and establish a baseline. In this work, we show the assisting capabilities of the Hindi and Marathi languages for the NER task. We show that models trained using multiple languages perform better than a single language. However, we also observe that blind mixing of all datasets doesn't necessarily provide improvements and data selection methods may be required. 
\end{abstract}

\renewcommand\IEEEkeywordsname{Keywords}
\let\thefootnote\relax\footnotetext{\textsuperscript{*} Equal contribution of the authors.}
\section{Introduction}
Named Entity Recognition \cite{8629225} is a popular information extraction approach used in natural language processing. It was introduced in 1995. This technique comprises detecting a named entity and then classifying it. These entities include names of locations, persons, organizations, and numeric values like financial values, time, and dates. 

The process of Named Entity Recognition (NER) is performed using three major techniques \cite{9039685}. These consist of rule-based NER, machine-learning-based NER, and hybrid NER. Rule-based NER is a process consisting of rules formalized by linguists. It includes a gazetteer, lexicalized grammar, etc. Machine-learning-based NER is a method in which models are trained using annotated text. Various machine-learning-based techniques \cite{inproceedings11} include Hidden Markov Model (HMM), Conditional Random Fields (CRFs), etc. Subsequently, deep learning architectures have also been explored for NER. Some examples of these architectures include Convolutional Neural Network (CNN) \cite{8308186}, Long Short Term Memory (LSTM) \cite{article11}, and Bi-Directional Long Short Term Memory (Bi-LSTM) \cite{Rrubaa}. Hybrid NER uses a combination of both of these methodologies.
 
Previously, a lot of work has been done on NER in the English language, owing to which a large number of resources are available for the English language. Indian languages have a small amount of NER literature. This is mainly because of data sparsity and a lack of tools owing to the intricacies of the languages. These intricacies make it difficult for the existing algorithms to be applied to low-resource languages \cite{joshi2022l3cube_mahanlp}. 

In order to overcome these difficulties and compensate for the lack of data, cross-lingual transfer training has been utilized in the domain of named entity recognition. This technique focuses on conserving and transferring knowledge gained while solving one problem to a distinct but related problem. It consists of training a model on large corpora in one language and sharing that knowledge with a model trained on a small amount of data in a different language. This makes it easier to enhance the entity classification of low-resource languages by taking advantage of the data from different languages. It is commonly known as low-resource cross-lingual transfer learning \cite{schuster-etal-2019-cross-lingual}. 
\begin{figure}[h!]
  
  \frame{\includegraphics [scale= 0.35] {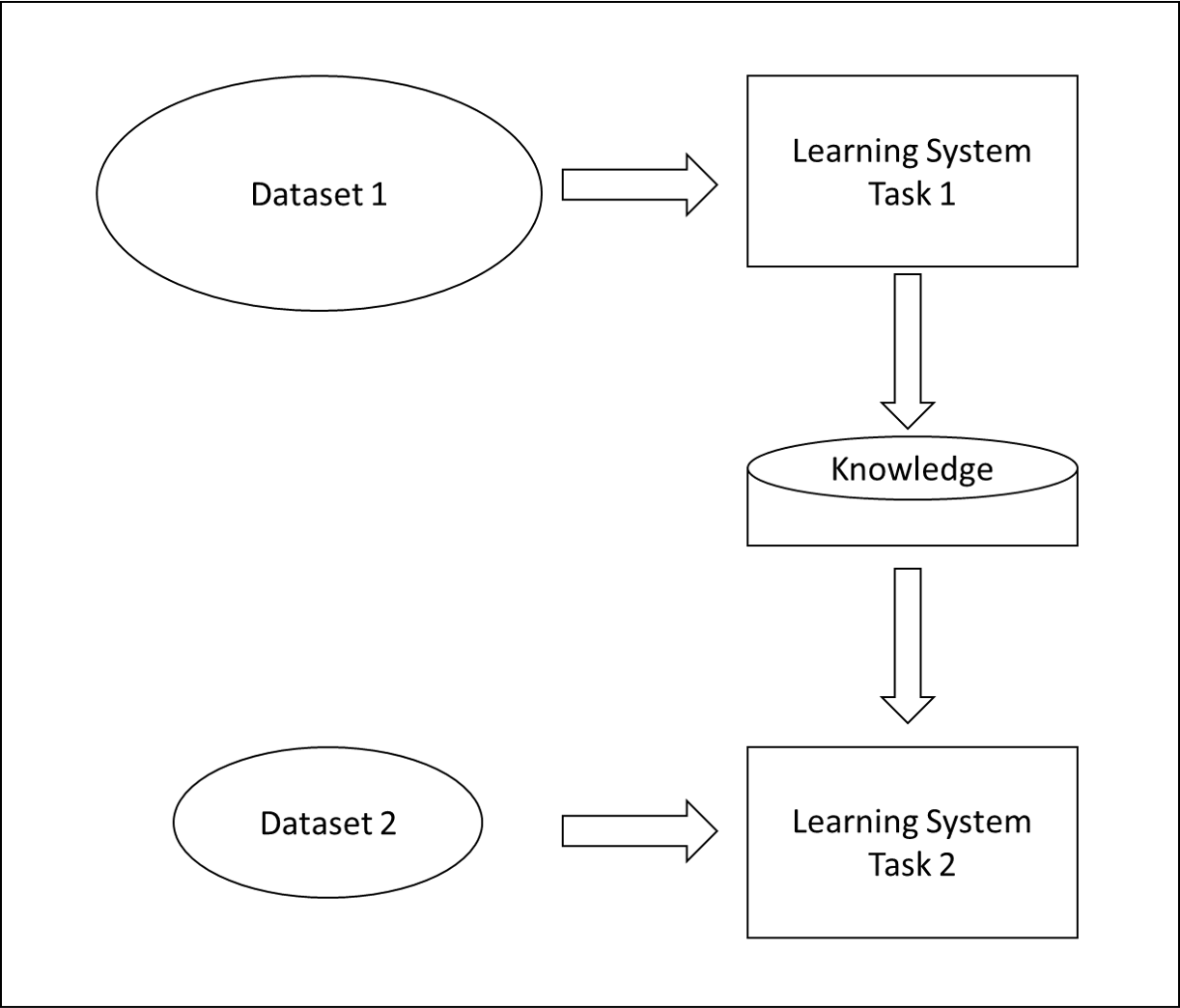}}
  \centering\caption{Model Architecture}
  \label{fig:1}
\end{figure}
The motivation was that very less data is available for low-resource Indian languages. We are establishing a benchmark for a low-resource Indian language that has significantly less amount of data and assist it with transfer learning with a similar low-resource language which can help increase the performance of the model.
In this paper, we have evaluated multi-lingual BERT models trained on Hindi and Marathi datasets and compared these results with their monolingual counterparts \cite{joshi2022l3cube}. These two languages follow the same script (Devanagari) and have similar grammatical rules and significant overlapping words. This motivated us to perform the cross-lingual evaluation using these languages and benchmark their assisting capabilities. The key contribution of this work is as follows:
\begin{itemize}
    \item We offer a detailed study of the architectures for transfer learning NER.
    \item We compare the results of deep learning models trained using traditional monolingual methods and multilingual methods. 
    \item We portray the advantages of cross-lingual transfer learning and its assistance in improving the results for Hindi and Marathi languages. 
\end{itemize}

The remainder of the paper is laid out as follows. Section 2 examines progress in Named Entity Recognition, with a concentration on Indian languages. The third section describes how we set up our experiments to examine different models. The results of all of the tests are summarised in Section 4. The conclusion of our paper is stated in Section 5.

\section{Related Work}
\cite{grishman-sundheim-1996-message} in 1995 introduced the concept of Named Entity Recognition at Message Understanding Conferences (MUC) in the US. The study was widespread for Indian languages around 2008. A study was carried out using various traditional machine learning algorithms and deep learning techniques. A comparative study between Support Vector Machine (SVM) and Conditional Random Field(CRF) trained on the same data was carried out by \cite{Krishnarao2009ACS}. The CRF model turned out to be the superior one. Hybrid systems consisting of a combination of the Hidden Markov Model, MaxEnt and handmade rules were found to be more accurate while performing NER as per \cite{srihari-2000-hybrid}. Deep learning models were then utilized to perform the NER problem as technology progressed. \cite{8308186}Convolutional Neural Network (CNN)(d), \cite{10.1162/neco.1997.9.8.1735}, \cite{9301306} and \cite{arkhipov-etal-2019-tuning} were among the most popular models used for NER.

\cite{article} describes various NER techniques used for Indian languages so far. The study examines the strategies for identifying named entities and compares them. It compares the Hidden Markov Model (HMM) with the Conditional Random Field (CRF) methods for Indian languages and concludes that the CRF method is the most effective. Work on NER for the Hindi language is carried out by \cite{Bhattacharjee2019NamedER} Machine Learning (ML), Rule-based, and Hybrid approaches are compared. The present NER systems are trained to operate on specified datasets and do not produce results on universal datasets. Hence, the study tries to uncover the shortcomings in the existing NER systems, particularly in the Hindi language. The article concludes that while the machine learning approach is more systematic when predicting unknown entities, it has a lesser accuracy than the rule-based system. Obstacles and concerns regarding NER for Marathi were discussed by \cite{a} along with its importance. It also examines several strategies and techniques for developing learning resources required for extracting NEs from unstructured natural language data.

\cite{murthy-etal-2018-judicious} shows how changes in tag distributions of commonly occurring named entities across primary and assisting languages affect the multilingual learning efficacy. To overcome this problem, the research presents a measure based on symmetric KL divergence that uses neural networks such as CNNs and Bi-LSTMs to filter out highly divergent training samples in the aiding language.

\cite{unknown} established baseline numbers for various publicly available Hindi and Marathi datasets. \cite{a} introduces a publicly available golden dataset L3Cube-MahaNER for the Marathi language containing 25000 sentences. \cite{l} in his paper addresses the problem of data sparsity by stealing traits from a closely comparable language's data. Hierarchical neural networks are used to train a supervised NER system. The network's common layers are used to borrow a feature from a closely comparable language. Multilingual Learning is when a neural network is trained on a combined dataset of a low-resource language and a closely related language. Our work is similar to theirs as we combine various datasets for Marathi and Hindi languages and compare them against mono-lingual models.

\section{Experimental Setup}
\begin{table}[htbp]
\caption{Count of individual tags of IIT Bombay corpora}
\centering
\begin{tabular}{llll}
\hline
\textbf{Tags} & \textbf{Train} & \textbf{Test} & \textbf{Validation}\\
\hline
{O} & {61235} & {28215} & {7349} \\
{NEL} & {4821} & {3148} & {796} \\
{NEP} & {1546} & {760} & {194} \\
{NEO} & {173} & {91} & {31} \\
\hline
\end{tabular}
\begin{tabular}{lc}
\hline

\end{tabular}
\end{table}

\begin{table}[htbp]
\caption{Count of individual tags of IJCNLP}
\centering
\begin{tabular}{llll}
\hline
\textbf{Tags} & \textbf{Train} & \textbf{Test} & \textbf{Validation}\\
\hline
{O} & {199481} & {42655} & {42068} \\
{NEL} & {3257} & {692} & {880} \\
{NEP} & {3893} & {886} & {806} \\
{NEO} & {2119} & {459} & {392} \\
\hline
\end{tabular}
\begin{tabular}{lc}
\hline

\end{tabular}
\end{table}

\begin{table}[htbp]
\caption{Count of individual tags of Wiki ANN Marathi}
\centering
\begin{tabular}{llll}
\hline
\textbf{Tags} & \textbf{Train} & \textbf{Test}\\
\hline
{O} & {46011} & {19633} \\
{NEL} & {8711} & {3958} \\
{NEP} & {1115} & {4671} \\
{NEO} & {10129} & {4310} \\
\hline
\end{tabular}
\begin{tabular}{lc}
\hline
\end{tabular}
\end{table}

\begin{table}[htbp]
\caption{Count of individual tags of Wiki ANN Hindi}
\centering
\begin{tabular}{llll}
\hline
\textbf{Tags} & \textbf{Train} & \textbf{Test}\\
\hline
{O} & {20015} & {9243} \\
{NEL} & {4469} & {2768} \\
{NEP} & {13015} & {2678} \\
{NEO} & {11102} & {6150} \\
\hline
\end{tabular}
\begin{tabular}{lc}
\hline
\end{tabular}
\end{table}

\begin{table*}
\centering\caption{Count Of sentences and tags in the datasets}
\centering
  \begin{tabular}{|l|l|l|l|l|l|l|}
    \hline
    Dataset &
      \multicolumn{3}{c}{\quad Count of Sentences } &
      \multicolumn{3}{c|}{\quad Count of Tags}\\
      
    & \quad Train \quad\quad & \quad Test \quad\quad  &\quad  Tune  \quad\quad&\quad  Train \quad\quad & \quad Test \quad\quad &\quad  Tune\quad\quad  \\
    \hline
    
    IJCNLP 200 NER Corpus &\quad 7979 & \quad 1711 & \quad 1710 &\quad  208750 & \quad 44692\quad  &\quad 44146 \\
    \hline
   IIT Bombay Marathi NER Corpus \qquad& \quad 3588 & \quad 1533 & \quad 470 &\quad  67775 & \quad 32214 &\quad 8370 \\
    \hline
    WikiAnn NER Corpus(Marathi)  &\quad  10674 &\quad  4304 & \quad -- & \quad 76006 & \quad 32572 &\quad\quad --  \\
    \hline
    WikiAnn NER Corpus(Hindi) & \quad 8356 &\quad  3477 &\quad-- &\quad 48601 &\quad  20829 &\quad\quad --\\
    \hline
  \end{tabular}
\label{table:1}
\end{table*}

\begin{table*}
\caption{\label{result-tab} F1 scores for the models tested on merged Hindi \& Marathi datasets and individual dataset. \\ IJCNLP(H) = IJCNLP Hindi Dataset, IITB(M) = IITB Marathi Dataset, Wiki(H) = WikiANN Hindi, \\ Wiki(M) = WikiANN Marathi}

\begin{center}
\begin{tabular}{{p{1.5cm}p{1.5cm}p{1.3cm}p{0.7cm}p{1.3cm}p{1cm}p{1.3cm}p{1cm}p{1.3cm}}}
\hline \textbf{Test Datasets} & \textbf{Train Datasets} & \textbf{Mbert} & \textbf{Indic Bert} & \textbf{Xlm Roberta} & \textbf{Maha Albert} & \textbf{Roberta Hindi} & \textbf{Maha Bert} & \textbf{Maha Roberta} \\
\hline
    IITB Marathi & IJCNLP(H) \&IITB(M) & 60.9 & 63.12 & \textbf{64.36} & 63.05 & \textbf{45.05} & \textbf{65.29} & \textbf{64.54}\\ \cline{2-9}
    
    &IITB(M) & 61.45 & 63.26 & 60.97 & 63.43 & 43.27 & 62.07 & 64.18  \\
    \hline
    
    IJCNLP Hindi& IJCNLP(H)  \&IITB(M) & 72.7 & \textbf{76.85} & \textbf{80.34} & \textbf{75.61} & \textbf{70.02} & \textbf{76.9} & \textbf{81.3}\\ \cline{2-9}
    
    &IJCNLP(H) & 75.52 & 76.32 & 79.71 &  63.11 & 70.01 &  76.53 &  78.67  \\
    \hline
    
    WikiANN Marathi& Wiki(H\&M) & \textbf{87.3} & 86.86 & \textbf{87.47} & 86.33 & \textbf{83.52} & \textbf{88.58} & 87.94\\ \cline{2-9}
    
    &Wiki(M) & 86.49 & 87.03 & 87.38 7 & 87.15 & 82.5 & 88.18 & 88.9 \\
    \hline
    
    WikiANN Hindi& Wiki(H\&M) & \textbf{83.85} & \textbf{84.26} & \textbf{85.94} & \textbf{83.33} & \textbf{82.88} & \textbf{82.83} & \textbf{85.9}\\ \cline{2-9}
    
    &Wiki(H) & 81.21 & 82.65 & 83.04 & 81.68 & 80.52 & 81.95 & 80.66  \\
    \hline
    
\end{tabular}
\end{center}
\end{table*}

\begin{table*}
\caption{\label{result-tab} F1 scores for the models tested by merging all datasets and individual dataset}
\begin{center}
\begin{tabular}{{p{2.3cm}p{2cm}p{2cm}p{2cm}p{2cm}p{2cm}}}
\hline \textbf{Model Name} & \textbf{Dataset type} &  \textbf{IIT Bombay} & \textbf{IJCNLP Hindi} & \textbf{WikiANN Marathi} & \textbf{WikiANN Hindi} \\
\hline
    
    mbert & Merged& 57.59 & 74.83 & 85.48 & \textbf{83.85}\\ \cline{2-6}
    
    & Mono & 61.45& 75.52 & 86.49 & 81.21\\
    \hline
    
    Indic Bert & Merged & 59.62 & 74.64 & 86.33 & \textbf{83.59}\\ \cline{2-6}
    
    & Mono & 63.26 & 76.32 & 87.03 & 82.65\\
    \hline
    
    Xlm Roberta & Merged& 60.73 & 78.58 & 86.78 & 84.21\\ \cline{2-6}
    
    & Mono & 60.97 & 79.71 & 87.38 & 83.04\\
    \hline
    
    MahaBert & Merged&  \textbf{62.76} & 74.98 & 86.98 & \textbf{83.19} \\ \cline{2-6}
    
    & Mono & 62.07 & 76.53 & 88.18 & 81.95\\
    \hline
    
    MahaRoberta & Merged& 64.11 & \textbf{79.22} & 87.17 & \textbf{84.36} \\ \cline{2-6}
    
    & Mono & 64.18 & 78.67 & 88.9 & 80.66\\
    \hline
    
    Maha Albert & Merged& 58.12 & \textbf{72.38} & 85.64 & 79.86 \\ \cline{2-6}
    
    & Mono & 63.43 & 63.11 & 87.15 & 81.68 \\
    \hline
    
    Roberta Hindi & Merged& \textbf{44.65} & 68.72 & \textbf{83.01} & \textbf{82.11} \\ \cline{2-6}
    
    & Mono & 43.27 & 70.01 & 82.5 & 80.52\\
    \hline

\end{tabular}
\end{center}
\end{table*}
We are confining our work on Named Entity Recognition (NER) to Marathi and Hindi, two of the top three languages spoken in India, and performed Named Entity Recognition on all publicly available datasets.

We are utilizing the dataset published in IJCNLP in 2008 and the WikiAnn NER Corpus published by \cite{pan-etal-2017-cross} in 2017 for the Hindi language. The IJCNLP dataset has 11,400 sentences in total. There are 12 categories in the IJCNLP dataset: organization, person, title-object, number, abbreviation, location, brand, title-person, time, measurement, designation, and words. We divided the data into 70-15-15 train, test, and tune segments because no data split was supplied. \cite{pan-etal-2017-cross} dataset for Hindi which is a silver standard dataset consisting of three categories: Location, Person, and Organization containing a total of  11,833 sentences. We are utilizing the IIT Bombay Marathi NER dataset provided by \cite{murthy-etal-2018-judicious} in 2018 and the WikiAnn NER Corpus dataset given by \cite{pan-etal-2017-cross} in 2017 for the Marathi language. There are a total of 5,591 sentences in the IIT B dataset. It is divided into three categories: location, person, and organization. The dataset is already separated into train-test-tune segments. \cite{pan-etal-2017-cross} dataset for Marathi has a total of 14,978 phrases grouped into three categories: Organization, Person, and Location. Again, it is a silver-standard dataset. Both datasets are in the IOB format.

The difficulties encountered while dealing with these datasets were as follows. Both datasets consisted of English terms. In the IJCNLP dataset, tagging was non-uniform, and the dataset included several confusing tags. In the IIT Bombay Marathi NER Corpus, more than 39 percent of sentences contained O tags, and in IJCNLP Corpus, more than 68 percent of the sentences contained O tags.

Several Marathi and Hindi datasets are available for NER, however, they are not publicly accessible. These datasets names are as follows: a) FIRE-2013- Indian Languages with Named Entity Recognition b) FIRE 2014 - Indian Languages with Named Entities. c) FIRE 2015- Entity Extraction from Indian Language Social Media Text (ESM-IL) d) FIRE 2016 is a joint task on code mix entity extraction in Indian languages (CMEE-IL) e) Marathi TDILNamed Entity Annotated Corpora. f) TDIL has been designated as the Entity Corpora for Hindi, Marathi, and Punjabi.

For uniformity and consistency in training the models, we have considered only location(NEL), person(NEP), and organisation(NEO) as tags. For the IJCNLP dataset we have considered only 'NEP' 'NEO' 'NEL' as tags and the remaining tags 'NETI', 'NETE', 'NEA', 'NED', 'NEM', 'NEN', 'NETO' were replaced with ‘O’. For the IIT Bombay NER dataset, we have removed the IOB notation and converted them into non-IOB format for example, 'B-PERSON' 'I-PERSON' has been converted to 'NEP' 'NEP' and 'B-ORGANISATION' 'I-ORGANISATION' to 'NEO' 'NEO' and 'B-LOCATION' 'I-LOCATION' to 'NEL' 'NEL'. Similarly, for the wiki-ann Marathi dataset 'B-ORG' 'I-ORG' to 'NEO' 'NEO' and 'B-PER' 'I-PER' to 'NEP' 'NEP' and 'B-LOC' 'I-LOC' to 'NEL' 'NEL'.
For the Wiki-ann Hindi dataset, 'B-ORG' 'I-ORG' to 'NEO' 'NEO' and 'B-PER' 'I-PER' to 'NEP' 'NEP' and 'B-LOC' 'I-LOC' to 'NEL' 'NEL'.
We have merged the IJCNLP train set and the IIT Bombay Marathi train set, the IJCNLP tune set and IIT Bombay Marathi tune set, and then reshuffled the sentences and evaluated the whole on various transformer-based models like mBert, Indicbert, and XlmRoBERTa, etc by testing it individually on the IIT Bombay test set and IJCNLP test set. Similarly, for the Wiki-Ann corpus, we have merged the Wiki-Ann Hindi train set and Wiki-Ann Marathi train set and evaluated models by testing individually on the Wiki-Ann Marathi test set \& Wiki-Ann Hindi test set. In the end, we merged all four datasets i.e., IJCNLP, IIT Bombay Marathi dataset, Wiki-Ann Marathi, and Wiki-Ann Hindi, and evaluated the results individually on each test set.

\section{Experimental Techniques}
\subsection{Model Architecture}
Transformers are a type of deep learning model that can process sequential data, such as natural language, by learning representations of the data that capture its underlying structure. 
In a transformer-based NER model, the transformer architecture is used to encode the input text into a sequence of hidden representations, which are then fed into a classifier to predict the named entities. The context is detected by the transformer's ability to capture the relationships between the words in the input text, which is learned through a self-attention mechanism.

Self-attention allows the transformer to attend to different parts of the input sequence and assign different weights to them based on their relevance to the task at hand. This means that the transformer can learn to recognize the context in which named entities appear, such as whether they are part of a person's name, the name of an organization, or a location.

Once the transformer has learned to recognize the relevant context for named entities, the classifier can use this information to make more accurate predictions. This process is typically trained on a large dataset of labeled examples, where the model learns to identify the context of named entities from the input text and associate them with the correct label.
The transformer in natural language processing attempts to handle sequence-to-sequence problems while simultaneously addressing long-range relationships. An "attention" mechanism in transformers helps to examine a connection between all of the words in a sentence. This aids in establishing the process of word evaluation to show which phrase components are most important to offer differential weightings and thus explains why ambiguous elements are solved quickly and efficiently. For example, if a sentence is provided as input, the transformer detects the context that provides the meaning of each word in the statement. The training time decreases as the feature improves parallelization. 

\textbf{BERT:}  BERT \cite{devlin2019bert} is a transformer-based approach created by Google for NLP pre-training referred to as Bidirectional Encoder Representations from Transformers. BERT learns about both sides of a token's context while training. Using unlabeled text by focusing on both sides of context left and right simultaneously across all layers, BERT  pre-trains deep bidirectional representations. To provide good-performing models for a vast range of tasks BERT model may be fine-tuned with just one extra output layer

\textbf{mBERT:} mBERT \cite{mbert}, referred to as multilingual BERT has been trained on and can be used in 104 languages. When mBERT was built, data from all 104 languages were combined. As a result, at the same time, mBERT incorporates and comprehends word connections in all 104 languages.

\textbf{ALBERT:} A Lite Bert (ALBERT) \cite{albert} is a Google AI deep-learning natural language processing (NLP) model that uses fewer parameters while retaining accuracy. When compared to BERT models, these models have better data throughput and can train approximately 1.7 times more quickly. On a large-scale corpus of 12 main Indian languages multilingual ALBERT model is trained which is referred to as IndicBERT. As compared to other multilingual models like mBERT and  XLM-R, IndicBERT contains many fewer parameters yet it achieves comparable or superior performance to other models.
\textbf{RoBERTa:}  RoBERTa \cite{roberta} is trained on a large corpus of English data which is a self-supervised model which was pre-trained on raw texts solely. With no human labeling, using an automated procedure RoBERTa produces inputs and labels from those texts. It was trained primarily with the Masked language modeling (MLM) objective. XLM-RoBERTa is a RoBERTa multilingual variant. It has been pre-trained on 2.5TB of filtered CommonCrawl data from 100 languages.

\section{Results}
We now discuss our results for the multilingual models trained on both Hindi and Marathi NER datasets. The aim is to evaluate assisting capability of these languages in improving individual language performance.

Initially, we merged the IIT Bombay(Marathi) with IJCNLP(Hindi) for training and tested it individually with the IIT Bombay and IJCNLP datasets. We then compared it with the results obtained on models trained with single monolingual data. The observations portray that XLM Roberta, RoBERTa Hindi, MahaBERT \cite{joshi2022l3cube}, and Maha Roberta perform better on the mixed dataset than on the monolingual IIT Bombay dataset (Marathi). Similarly, when the mixed data model is tested on IJCNLP (Hindi), the scores observed are better for all the models except with mBERT where monolingual IJCNLP performs better. 

Likewise, Wiki ANN datasets for Hindi and Marathi were merged and tested with individual datasets to compare the results with models that were trained and tested on monolingual data. Again the majority of the models, when tested with individual datasets, outperformed the models trained on monolingual datasets of Wiki ANN.

Lastly, we merged all the datasets, viz. IIT Bombay(Marathi), IJCNLP(Hindi), and Wiki ANN(Hindi and Marathi) and tested them individually on the merged dataset. We see consistent improvements in Wiki ANN(Hindi) dataset using this expanded training set. However, the improvements are not consistently visible for other datasets. This shows that blind mixing of all the available datasets may not be desirable, and some data selection methods may be further employed. The difference in the domain of the Wiki ANN dataset and other datasets is possibly responsible for this behavior. The difference in the domain of the Wiki ANN dataset and other datasets is possibly responsible for this behavior and needs further investigation. 
NER Models performance can be done by increasing the number and quality of training data. To understand patterns and relationships between words and named items, NER models rely significantly on training data. Increasing the number and quality of the training data can aid the model in capturing the nuances and changes of named entities in various situations. Using pre-trained language models. Language models such as BERT, GPT-2, and others have been demonstrated to improve NER performance. These models are trained on enormous amounts of data and can capture linguistic and contextual nuances, which can help improve NER performance. Use of Fine-tune pre-trained models can be used by adjusting them to the unique task and domain, fine-tuning pre-trained language models on specific NER tasks can help to improve their performance. Incorporating external knowledge sources, such as gazetteers, dictionaries, and ontologies, can help to improve NER performance by providing additional information on named entities that may not be included in the training data. Ensemble learning can be accomplished by training numerous models on distinct subsets of the training data, or by mixing models of various sorts, such as rule-based models, statistical models, and neural network models.

\section{Conclusion}

The fundamental goal of this work is to develop deep learning-based multilingual models for named entity identification in Indian languages. We aim to benchmark multilingual transformer-based models on Hindi and Marathi NER datasets. We also explore the assisting capabilities of these languages. We show that combining the two languages does provide improvements in individual languages. However, mixing all the available datasets results in slight degradation of the performance on most of the individual datasets.

\section*{Acknowledgements}
This work was done under the L3Cube Pune mentorship
program. This work is a part of the L3Cube-MahaNLP\cite{joshi2022l3cube_mahanlp} project, the problem statement and ideas presented in this work originated from L3Cube and its mentors. We would like to express our gratitude towards
our mentors at L3Cube for their continuous support and
encouragement.

\vspace{12pt}
\bibliography{main.bib}
\bibliographystyle{IEEEtran}

\end{document}